\documentclass{article}
\usepackage{spconf,amsmath,graphicx,hyperref,enumitem, multirow, booktabs, makecell}

\title{Learning from Disagreement: A Group Decision Simulation Framework for Robust Medical Image Segmentation}
\name{Chen Zhong$^{1}$, Yuxuan Yang$^{1}$, Xinyue Zhang$^{2}$, Ruohan Ma$^{2}$, Yong Guo$^1$, Gang Li$^2$, Jupeng Li$^{1*}$
\thanks{$^{*}$Corresponding author: lijupeng@bjtu.edu.cn}}

\address{
    $^1$School of Electronics and Information Engineering, Beijing Jiaotong University, China \\
    $^2$Peking University School and Hospital of Stomatology, China
}
\begin{document}
\ninept
\maketitle
\begin{abstract}
Medical image segmentation annotation suffers from inter-rater variability (IRV) due to differences in annotators' expertise and the inherent blurriness of medical images. Standard approaches that simply average expert labels are flawed, as they discard the valuable clinical uncertainty revealed in disagreements. We introduce a fundamentally new approach with our group decision simulation framework, which works by mimicking the collaborative decision-making process of a clinical panel. Under this framework, an Expert Signature Generator (ESG) learns to represent individual annotator styles in a unique latent space. A Simulated Consultation Module (SCM) then intelligently generates the final segmentation by sampling from this space. This method achieved state-of-the-art results on challenging CBCT and MRI datasets (92.11\% and 90.72\% Dice scores). By treating expert disagreement as a useful signal instead of noise, our work provides a clear path toward more robust and trustworthy AI systems for healthcare.
\end{abstract}
\begin{keywords}
Probablistic Segmentation, Medical Images, Image Segmentation, Inter-rater Variability, Uncertainty Estimation
\end{keywords}
\section{Introduction}
\label{sec:intro}

Medical image segmentation is a fundamental task in medical image analysis \cite{isensee2018nnu}. Over the last decade, deep learning (DL) models, including convolutional neural networks (CNNs) and vision transformers (ViT) approaches \cite{kofler2021we}, have been developed for image segmentation. Such supervised deep learning segmentation algorithms rely on the assumption that the provided reference annotations for training reflect the unequivocal ground truth. Yet, in many cases, the labeling process contains substantial inconsistencies in annotations manifest from differences in annotators' expertise and the inherent blurriness of medical images \cite{antonelli2022medical}. Therefore, uncertainty estimation has rapidly developed in medical image segmentation, which is also referred to probabilistic segmentation, preventing erroneous impacts on subsequent analysis and diagnosis.  

One of the most thoroughly investigated methods for uncertainty estimation in DL is with Bayesian framework \cite{luo2023efficient, armato2011lung, jha2019kvasir}. In the Bayesian framework, prior knowledge (e.g. neural network weights) are combined with data via Bayes theorem. While Bayesian-based methods are supported by rich theory for quantifying both aleatoric and epistemic uncertainty, it often entail considerable computational expenditure. To address this, the flexible and easily implementable approach is Monte Carlo dropout, which involves randomly removing elements of the input tensor while using spatial dropouts. However, this serves somewhat crude approximation to the posterior. More recently, Variational inference has emerged as a competitive tool for probabilistic segmentation for fast approximate Bayesian computations \cite{li2022learning, guo2022modeling, liao2024modeling}. Variational autoencoder and variational inference methods, such as the Probabilistic U-Net, employ a simpler parametric distribution to approximate the posterior to drastically simplify computations. However, these methods are limited by their reliance on a single latent space, which mixes together the actual image content with the unique style of each annotator \cite{guan2018said, liu2020ms, chen2019automatic, jensen2019improving, mirikharaji2019learning}. Because the model cannot separate these two sources of information, it is impossible to learn from individual expert decisions \cite{ji2021learning, liao2024modeling, yu2020difficulty}. As a result, the rich and diagnostically valuable disagreements between experts are incorrectly dismissed as random noise.

To these challenges, we propose a novel probabilistic segmentation method that employs Pyramid Vision Transformer (PVT) and variational autoencoder. Specifically, during the training phase, our Expert Signature Generator (ESG) constructs a latent space to model annotation styles. Through a hierarchical architecture, the ESE explicitly disentangles systematic expert 'signatures' from stochastic random errors and integrates these learned styles with image features from a pre-trained model. Following this, our Simulated Consultation Module (SCM) generates the final segmentation by simulating a clinical consultation. This process occurs during the sampling stage, where the SCM synthesizes diverse expert signatures from the latent space with semantic image features to produce the final output. The main contributions of this study are as follows:
\begin{itemize}[nosep]
   \item \textbf{A Novel Group Decision Simulated (GDS) Framework.} We propose a new paradigm that treats expert disagreement as a valuable signal, not noise. Our framework simulates a clinical panel’s decision-making process to intelligently synthesize diverse annotations rather than simply averaging them, better reflecting real-world clinical practice.
   \item \textbf{A Specialized Architecture for Style Disentanglement and Fusion.} We introduce an Expert Signature Generator (ESG) to disentangle expert ‘signatures’ from random error, and a Simulated Consultation Module (SCM) that uses adaptive sampling to generate a final segmentation balancing expert consensus with clinically relevant uncertainty.
   \item \textbf{State-of-the-Art Performance Across Diverse Medical Datasets.} Extensive experiments on public and private MRI and CBCT datasets validate our method’s superiority. Our model excels in both regions of high agreement and ambiguous areas, demonstrating its significant improvements in robustness and real-world clinical applicability.
\end{itemize}  

\section{Method}

\label{sec:format}
\begin{figure}
    \centering
    \includegraphics[width=1.0\linewidth]{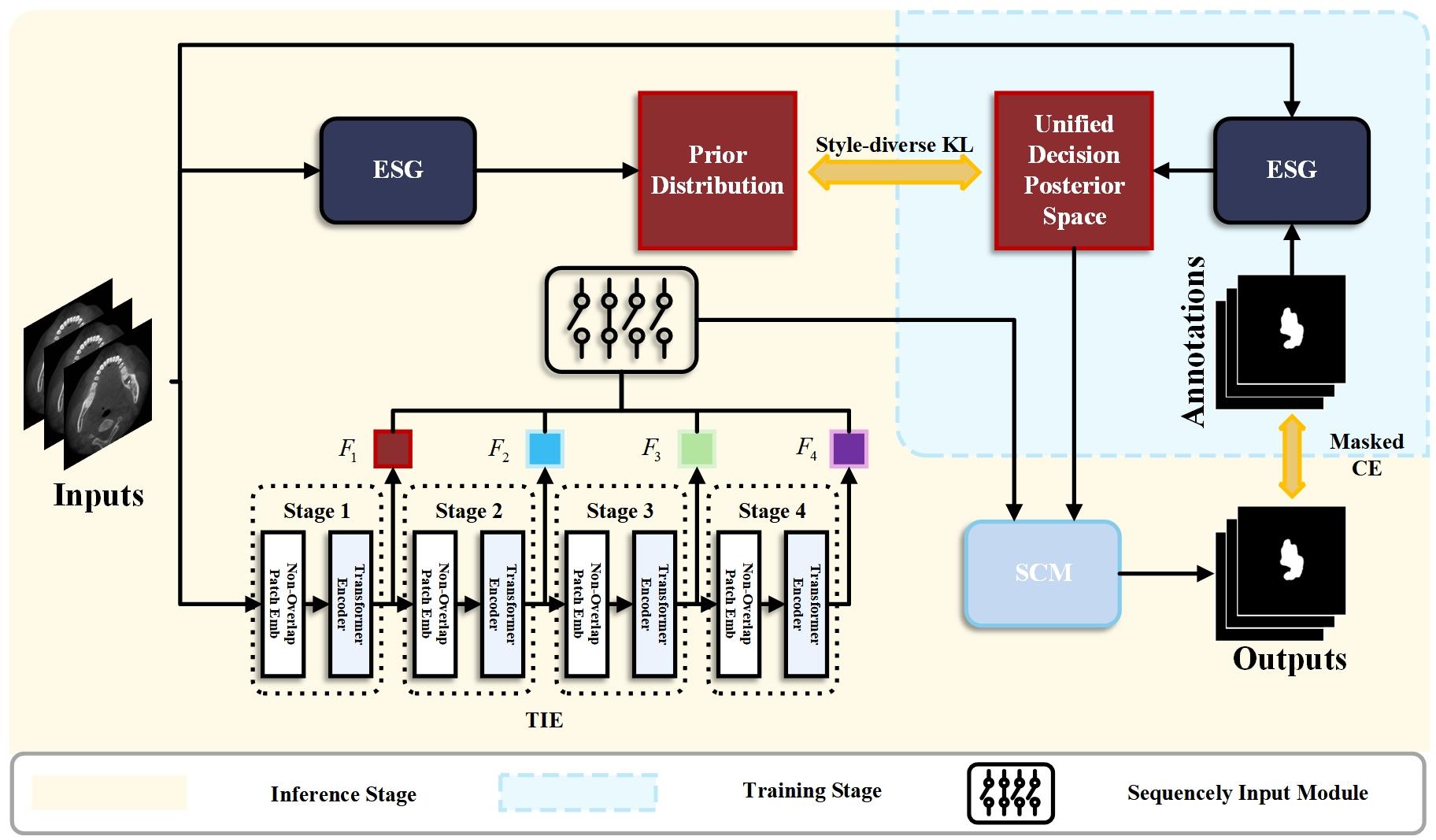}
    \caption{The overall structure of our proposed framework.}
    \label{fig:placeholder}
\end{figure}

The principled design of our GDS framework is motivated by a crucial insight into the composite nature of IRV. Our analysis reveals that IRV is not a single, monolithic source of noise, but rather stems from two distinct phenomena:  

First, we identify minor stochastic errors, which manifest as small, localized deviations (approx. 1–5 pixels) when a single clinician repeatedly annotates an image. These can be attributed to random noise or slight inconsistencies in manual delineation.  

Second, and more significantly, we identify systematic annotator-specific biases, which arise from differing diagnostic preferences among experts. These result in larger, structurally meaningful disparities (often exceeding 5 pixels) that represent valid, alternative clinical interpretations.  

Crucially, we postulate that these two sources of variability operate at different semantic scales. Annotator bias influences the overall shape and macro-level judgment, while stochastic error is a fine-grained, pixel-level perturbation. A model that fails to distinguish between them would incorrectly penalize meaningful clinical disagreements as if they were simple noise.

\subsection{Architecture Overview}
\label{sec:overview}

As depicted in Fig. 1, the proposed GDS framework consists of a PVT backbone, an Expert Signature Generator (ESG), and a Simulated Consultation Module (SCM). Network modules selectively participate in the training and inference phases. Specifically, during the training phase, for each input image $x$, the PVT backbone first extracts its feature map $\{F_{1}, F_{2}, F_{3}, F_{4}\}$. The core of our method involves two parallel applications of the ESG module to estimate a prior distribution based only on the image $x$, and a posterior distribution conditioned on both the image $x$ and annotation set $\{y_{1}, ..., y_{n}\}$. In this process, the posterior distribution learns to capture the unique style of an individual annotator. A latent code is then sampled from this posterior, and we define this sampled code as the expert signature $e_i$. With this established expert signature $e_i$, the SCD reconstructs a segmentation result. The framework is then optimized via a dual supervisory signal: a segmentation loss comparing the reconstruction with the corresponding annotation, and a KL divergence loss between the posterior and prior distributions. We now delve into the details of our GDS framework.

\subsection{Backbone}

We adopt the Pyramid Vision Transformer v2 (PVTv2) \cite{wang2021pyramid} as our feature extraction backbone, selected for its proven efficacy in capturing multi-scale representations. As a Transformer-based architecture, PVTv2 introduces a pyramid structure and spatial-reduction attention (SRA) to efficiently generate a feature pyramid, denoted as $\{F_{1}, F_{2}, F_{3}, F_{4}\}$. These multi-scale features provide the rich semantic and spatial context essential for our probabilistic segmentation task. To ensure a robust initialization, the backbone starts with ImageNet pre-trained weights, and its parameters are subsequently fine-tuned on our target datasets to better adapt to the specific domain.  

\subsection{Expert Signature Generator}  

To ensure the ESG learns a latent space that is explicitly conditioned on annotator identity, we introduce a style modulation mechanism, as depicted in Fig. 2. First, each annotator is represented by a one-hot vector, which is then projected into a high-dimensional annotator embedding using a lightweight DenseNet \cite{huang2017densely} encoder. This embedding functions as a style modulator, effectively acting as a learned signature that can steer the feature representation. The core of this mechanism is an element-wise multiplication between the annotator embedding and the feature maps extracted from the image-annotation pair. This operation modulates the features, infusing them with annotator-specific stylistic information. The resulting modulated features are then passed through subsequent convolutional layers to predict the parameters (mean and variance) of the posterior distribution, ensuring it is tailored to a specific expert's style. 

\subsection{Simulated Consultation Module}

Informed by our aforeheading finding that different types of IRV manifest at different feature scales, we designed an attention-guided multi-scale fusion strategy to intelligently inject the expert signature during the decoding stage.  

This process begins at each feature level $l$, where an expert signature $e_{i}$ is randomly sampled from the ESG generated posterior distribution, is spatially tiled and concatenated with the corresponding feature map $F_{l}$, from the PVTv2 backbone to produce a combined representation. Acknowledging that an expert's style may influence features differently across scales, we introduce a lightweight attention module to generate a dynamic attention map $\alpha_{l}$, which is formulated as follows:  

\begin{equation}
    \alpha_{l} = Atten(Concat(Tile(e_i), F_{l})
\end{equation}  

This map adaptively re-weights the combined features via element-wise multiplication, yielding the attention-aware feature $F_{l}'$. Finally, the set of attention-weighted features from all scales, $\{F_{1}', F_{2}', F_{3}', F_{4}'\}$, are upsampled to a uniform resolution, fused into a single representation and passed through a final convolutional layer to produce the ultimate segmentation prediction.

\section{Experiments and Result}

\subsection{Datasets and settings}

\textbf{Public:} QUBIQ 2021 Challenge \cite{li2024qubiq} is a publicly available dataset for IRV research that includes both 2D and 3D multi-modality segmentation tasks. Some datasets of this Challenge are omitted due to their tiny sizes, which makes evaluation extremely sensitive to a few test samples and inhibit meaningful analysis.  

\noindent\textbf{Private}: We collected the largest dataset to date for jaw cystic lesion segmentation, referred to as JCL, comprising 60 Cone Beam Computed Tomography (CBCT) scans and a total of 8,550 2D slices. Each scan has an in-plane resolution of 512×512 pixels, with pixel spacing ranging from 0.55 mm to 0.82 mm. The number of slices per volume ranges from 205 to 891, with slice thicknesses between 0.35 mm and 1.25 mm. All CBCT volumes were independently annotated by two board-certified oral and maxillofacial radiologists, each with over eight years of clinical experience. Data collection was conducted at Peking University School and Hospital of Stomatology, with Institutional Review Board (IRB) approval obtained from its Research Ethics Committee.

For each segmentation task, all images were normalized via subtracting the mean and dividing by the standard deviation on a pixel-by-pixel basis. For a fair comparison, we followed the setting in XXX for all public tasks.  All experiments were performed on a workstation with one NVIDIA RTX 3090 GPU and implemented under the PyTorch \cite{paszke2019pytorch} and MONAI \cite{cardoso2022monai} framework. Further details of the experiments setting are provided in Table 1.
The mean and standard deviation were counted on training cases. For a fair comparison, we followed the setting in \cite{ji2021learning}, i.e., setting the mini-batch size to 8 and resizing the input image to 256×256 for all public tasks. The images in the prostate subset in QUBIQ are relatively large, so we followed the setting in \cite{liao2024modeling}, i.e., center cropped to 640×640 before resizing to 256×256. The Adam optimizer \cite{kingma2014adam} with an initial learning rate of $lr_{0}$ = 1e-4 was adopted as the optimizer.

\begin{table}[]
    \caption{Hyperparameter settings for QUBIQ and JCL.}
    \centering
    \begin{tabular}{lccc}
        
        \toprule
        \multirow{2}{*}{\makecell[c]{\textbf{Hyper} \\ \textbf{Params}}} & \multicolumn{2}{c}{\textbf{QUBIQ}} & \multirow{2}{*}{\textbf{JCL}} \\ 
        \cline{2-3}
        & \textbf{Prostate} & \textbf{Kidney} & \\
        \midrule
        Base \(l_r\) & \(1 \times 10^{-4}\) & \(1 \times 10^{-4}\) & \(1 \times 10^{-4}\) \\
        Epoch & 120 & 120 & 150 \\
        Input size & 256×256 & 256×256 & 300×300 \\
        Batch size & 8 & 8 & 8 \\
        Optimizer & ADAM & ADAM & ADAM \\
        Weight decay & \(5 \times 10^{-4}\) & \(5 \times 10^{-4}\) & \(5 \times 10^{-4}\) \\
        Momentum & 0.9 & 0.9 & 0.9 \\ 
        \bottomrule
    \end{tabular}
    \label{table1}
\end{table}

\subsection{Comparison study}

\begin{table}[b]
    \centering
    \caption{Performance comparison of different methods in ambiguous and certain regions on JCL.}
    \label{tab:my_label} 
    \begin{tabular}{lcccc}
        \toprule
        \multirow{2}{*}{\textbf{Method}} & \multicolumn{2}{c}{\textbf{Ambiguous Regions}} & \multicolumn{2}{c}{\textbf{Certain Regions}} \\
        \cmidrule(lr){2-3} \cmidrule(lr){4-5}
        & GED$\downarrow$ & Dice$_{\text{soft}}\uparrow$ & GED$\uparrow$ & Dice$_{\text{soft}}\uparrow$ \\
        \midrule
        Prob U-Net & 0.7930 & 75.16 & 0.6901 & 74.73 \\
        D-Persona  & 0.5281 & 78.81 & 0.6545 & 76.98 \\
        AVAP       & 0.6530 & 80.16 & 0.7004 & 82.17 \\
        PADL       & 0.5001 & 82.23 & 0.6854 & 85.65 \\
        MH-PVT     & 0.7781 & 79.01 & \textbf{0.7881} & 73.12 \\
        Ours       & \textbf{0.3210} & \textbf{88.10} & 0.7715 & \textbf{90.01} \\
        \bottomrule
    \end{tabular}
\end{table}\textbf{}

\begin{table*}[]
    \centering
    \caption{Comparison of methods on ambiguous and certain regions.}
    \begin{tabular}{lcccccccc}
        \toprule
        & \multicolumn{4}{c}{\textbf{QUBIQ-prostate}} & \multicolumn{4}{c}{\textbf{QUBIQ-kidney}} \\
        \cline{2-9}
        \multirow{2}{*}{\textbf{Method}} & \multicolumn{2}{c}{\textbf{Ambiguous Regions}} & \multicolumn{2}{c}{\textbf{Certain Regions}} & \multicolumn{2}{c}{\textbf{Ambiguous Regions}} & \multicolumn{2}{c}{\textbf{Certain Regions}} \\
        & \textbf{GED$\downarrow$} & \textbf{Dice$_\text{soft}$$\uparrow$} & \textbf{GED$\uparrow$} & \textbf{Dice$_\text{soft}$$\uparrow$}  & \textbf{GED$\downarrow$} & \textbf{Dice$_\text{soft}$$\uparrow$} & \textbf{GED$\uparrow$} & \textbf{Dice$_\text{soft}$$\uparrow$}\\
        \midrule
        Prob U-Net & 0.2147 & 78.45 & 0.2541 & 75.18 & 0.1996 & 79.09 & 0.2492 & 77.10 \\
        D-Persona & 0.1784 & 84.12 & 0.1804 & 80.12 & 0.1586 & 86.45 & 0.1902 & 82.15 \\
        AVAP & 0.1975 & 86.14 & 0.1766 & 80.01 & 0.1884 & 88.51 & 0.1797 & 84.04 \\
        PADL & 0.2531 & 81.95 & 0.2009 & 84.32 & 0.2457 & 88.29 & 0.1790 & 86.78 \\
        MH-PVT & 0.2201 & 82.18 & 0.2391 & 86.01 & 0.2001 & 85.05 & 0.2058 & 87.18 \\
        \textbf{Ours} & \textbf{0.1655} & \textbf{90.72} & \textbf{0.3953} & \textbf{89.71} & \textbf{0.1403} & \textbf{91.02} & \textbf{0.3530} & \textbf{88.98} \\
        \bottomrule
    \end{tabular}
    \label{tab:table2}
\end{table*}

In medical image segmentation, label diversity is not inherently valuable unless it occurs in regions of diagnostic uncertainty. Generating diverse predictions in areas where expert consensus is high can reduce model reliability and hinder clinical trust. Therefore, meaningful diversity should be concentrated in regions with high inter-rater disagreement—areas where ambiguity is inherent and clinical decisions are more nuanced. Motivated by this insight, we analyze inter-rater entropy to identify ambiguous and certain regions within the annotated data. We then evaluate the model’s performance separately in these regions, aiming to better assess its ability to represent uncertainty in a clinically relevant and informative way.  

To better evaluate the diversity and clinical relevance of the generated labels, we first computed entropy maps based on the annotations from different raters. Using these maps, the annotated regions were divided into ambiguous regions and certain regions. We then calculated the Generalized Energy Distance (GED) and soft Dice score separately for each region. The results are presented in Table \ref{tab:table2}.  

Table \ref{tab:table2} compares the performance of several representative probabilistic segmentation methods across regions with varying levels of annotation certainty. The results reveal a clear trade-off in how different models handle ambiguous versus certain areas in medical image segmentation.  

In ambiguous regions, which correspond to clinically uncertain areas with notable inter-rater disagreement, our method achieves the lowest GED (0.1655) and the highest soft Dice score (90.72\%) among all compared approaches. This demonstrates that our model not only captures the diversity of expert opinions but also aligns well with the underlying distribution of annotations—offering both variability and consistency where it matters most. In contrast, methods such as PADL and MH-PVT report higher GED values (0.2531 and 0.2201, respectively), indicating less effective modeling of uncertainty, and are accompanied by lower Dice scores in these regions.  

In certain regions, where expert agreement is high and label diversity is minimal, our model still achieves excellent segmentation accuracy with a Dice score of 89.71\%, outperforming all other methods. Although the corresponding GED is 0.3953, which appears relatively high for such areas, this reflects the model’s deliberate preservation of subtle, possibly clinically relevant variations—even in regions of consensus. Rather than indicating noise or error, this controlled diversity may contribute to improved model robustness and adaptability. The high Dice score confirms that accuracy is not compromised in the process.  

Compared to baseline methods such as Prob U-Net and D-Persona, which tend to either overgeneralize or struggle to adapt across varying uncertainty levels, our method exhibits both discriminative uncertainty modeling and clinical reliability. For instance, while D-Persona achieves a competitive GED in ambiguous regions (0.1784), its significantly lower Dice score in certain regions (80.12\%) suggests weaker generalization to high-confidence areas. Conversely, PADL performs better in certain regions than in ambiguous ones, highlighting its limited ability to model uncertainty where it is most needed.  

On private dataset JCL with lower inter-rater variability, our method consistently achieves superior Dice scores in both ambiguous (88.10\%) and certain regions (90.01\%), demonstrating strong segmentation accuracy across different annotation confidence levels. While the GED values are higher compared to the public dataset (0.3210 vs. 0.1555 in ambiguous regions, and 0.7715 vs. 0.3953 in certain regions), this increase reflects the model’s deliberate preservation of prediction diversity—even in less ambiguous environments. This is particularly valuable in clinical settings where subtle anatomical variations still matter.  

Compared to other methods, our approach maintains a clear advantage: PADL and D-Persona exhibit trade-offs between accuracy and uncertainty handling, while Prob U-Net and MH-PVT consistently underperform across both region types. These findings highlight the robustness and adaptability of our method under different annotation conditions, suggesting it generalizes well to datasets with both high and low inter-rater variability.

\begin{figure}
    \centering
    \includegraphics[width=1.0\linewidth]{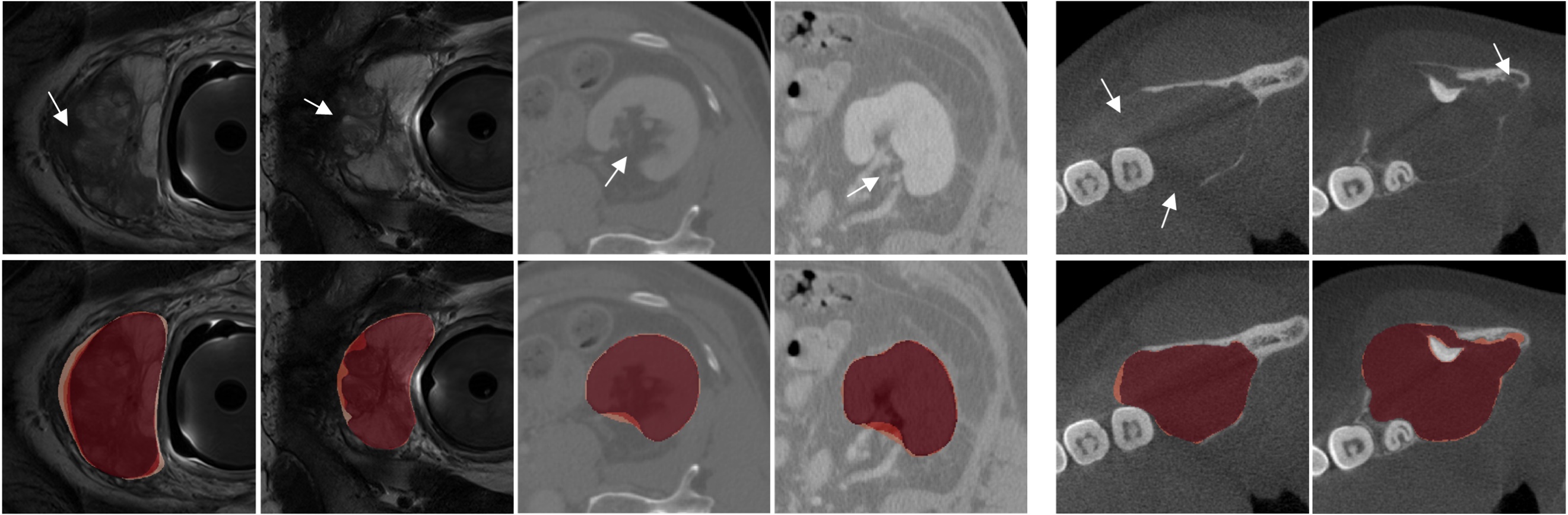}
    \caption{Diversified segmentation results of our proposed framework on the public QUBIQ (Left 4 column) and private JCL (Right 2 column) datasets. Different colors denote different delineations, which shows that our model only generate diverse and plausible predictions on inherent blurred areas in the medical image (white arrow).}
    \label{fig:placeholder}
\end{figure}

\subsection{Ablation study}

To evaluate the individual and combined contributions of the ESG and SCM components, we conducted ablation experiments on our private dataset characterized by relatively low inter-annotator variability. The results are summarized in the Table \ref{tab:ablation}.

\begin{table}[h]
    \centering
    \caption{Ablation study of GDS components on JCL dataset}
    \begin{tabular}{cccccc}
        \toprule
        \multirow{2}{*}{\textbf{ESG}} & \multirow{2}{*}{\textbf{SCM}} & \multicolumn{2}{c}{\textbf{Ambiguous Regions}} & \multicolumn{2}{c}{\textbf{Certain Regions}} \\
        \cline{3-6}
        & & \textbf{GED} $\downarrow$ & \textbf{Dice$_{soft}$} $\uparrow$ & \textbf{GED} $\uparrow$ & \textbf{Dice$_{soft}$} $\uparrow$ \\
        \midrule
        $\surd$ & & 0.3702 & 80.69 & 0.6513 & 84.12 \\
        & $\surd$ & 0.4180 & 81.04 & 0.7009 & 85.11 \\
        $\surd$ & $\surd$ & \textbf{0.3210} & \textbf{88.10} & \textbf{0.7715} & \textbf{90.01} \\
        \bottomrule
    \end{tabular}
    \label{tab:ablation}
\end{table}

ESG Alone (Row 1)
Introducing ESG significantly reduces the GED score in ambiguous regions (0.3702), indicating its strong ability to model rater-specific variability and improve the alignment with ground-truth uncertainty. However, its Dice scores, although improved compared to baseline methods, remain moderately lower than the full model.
SCM Alone (Row 2)
When only SCM is used, we observe a slightly better Dice score in both ambiguous (81.04) and certain (85.11) regions compared to ESG alone, suggesting that SCM contributes more directly to accurate segmentation by consolidating shared decision patterns across annotators. However, GED increases, especially in ambiguous regions (0.4180), implying reduced capability in capturing fine-grained annotation variability.
ESG + SCM (Full Model, Row 3)
When both components are integrated, we observe the best performance across all metrics. The model achieves the lowest GED in ambiguous regions (0.3210), suggesting accurate modeling of uncertainty, and simultaneously achieves the highest Dice scores in both ambiguous (88.10) and certain regions (90.01), confirming the effectiveness of the joint design.
These findings demonstrate the complementary strengths of ESG and SCM: ESG captures annotation diversity rooted in rater bias and image ambiguity, while SCM enhances global interpretability and decision consistency. Their combination results in a unified latent space that is both clinically aware and statistically robust, reinforcing the overall superiority of our framework.

\section{Conclusion and Discussion}

In this paper, we proposed a novel framework that simulates group decision-making processes to effectively address the pervasive challenge of inter-rater variability in medical image segmentation. Our approach centers on constructing a unified latent space that captures both semantic features of medical images and the unique interpretive tendencies of individual annotators. By jointly modeling these two aspects, our method—termed ESG—offers a more faithful representation of the annotation distribution, accommodating both consensus and disagreement in expert opinions.  

Theoretically, we demonstrate that annotation variability is inherently conditioned on both image content and annotator-specific factors. ESG explicitly disentangles and re-integrates these sources of variability, allowing the model to generalize across diverse annotation styles while preserving fine-grained clinical distinctions. Empirically, our framework achieves state-of-the-art performance across multiple benchmark datasets and modalities, outperforming existing probabilistic segmentation methods in both ambiguous and certain regions. Notably, it balances segmentation accuracy with uncertainty modeling, demonstrating superior Dice scores and lower Generalized Energy Distance (GED) values.  

Furthermore, extensive analysis shows that ESG not only handles high-variance regions with greater fidelity but also maintains strong generalization in high-consensus areas, preserving subtle but potentially meaningful variability. These findings underscore the clinical utility of our approach, particularly in scenarios requiring robust modeling of expert disagreement or generation of high-confidence consensus segmentations.  

Overall, this work provides a principled and practical solution to the longstanding problem of annotation noise in medical image analysis, laying the groundwork for more trustworthy, interpretable, and adaptable computer-aided diagnostic systems.  

\label{sec:refs}

\bibliographystyle{IEEEbib}
\bibliography{strings,refs}

\end{document}